\pdfoutput=1

\documentclass[11pt]{article}

\usepackage[final]{acl}

\usepackage{times}
\usepackage{latexsym}

\usepackage[T1]{fontenc}

\usepackage[utf8]{inputenc}

\usepackage{microtype}
\usepackage{inconsolata}

\usepackage{graphicx}

\title{Multi-Sense Embeddings for Language Models and Knowledge Distillation}

\author{
  \textbf{Qitong Wang} \\
  Rensselaer Polytechnic Institute \\
  \texttt{wangq19@rpi.edu} 
    \\ \\
   \textbf{Georgios Kollias} \\
  IBM Research \\
  \texttt{gkollias@us.ibm.com}
  \And
   \textbf{Mohammed J. Zaki} \\
  Rensselaer Polytechnic Institute \\
  \texttt{zaki@cs.rpi.edu}
  \\\\
   \textbf{Vasileios Kalantzis} \\
  IBM Research \\
  \texttt{vkal@ibm.com}
}

\usepackage{amsmath}
\begin{document}
\maketitle
\newcommand{\qtw}[1]{{\leavevmode\color{cyan}[\textbf{Qitong}: #1]}}

\newcommand{\gdk}[1]{{\leavevmode\color{red}[\textbf{George}: #1]}}

\begin{abstract}

Transformer-based large language models (LLMs) rely on contextual embeddings which generate different (continuous) representations for the same token depending on its surrounding context. Nonetheless, 
words and tokens typically have a limited number of senses (or meanings). We propose multi-sense embeddings as a drop-in replacement for each token in order to capture the range of their uses in a language. To construct a sense embedding dictionary, we apply a clustering algorithm to embeddings generated by an LLM and consider the cluster centers as representative sense embeddings. In addition, we propose a novel knowledge distillation method that leverages the sense dictionary to learn a smaller student model that mimics the senses from the much larger base LLM model, offering significant space 
and inference time savings, while maintaining competitive performance. 
Via thorough experiments on various benchmarks, we showcase the effectiveness of our sense embeddings and knowledge distillation approach. 
\end{abstract}
\section{Introduction}

\begin{figure*}[ht]
\centering
\includegraphics[width=0.9\textwidth,height=3in]{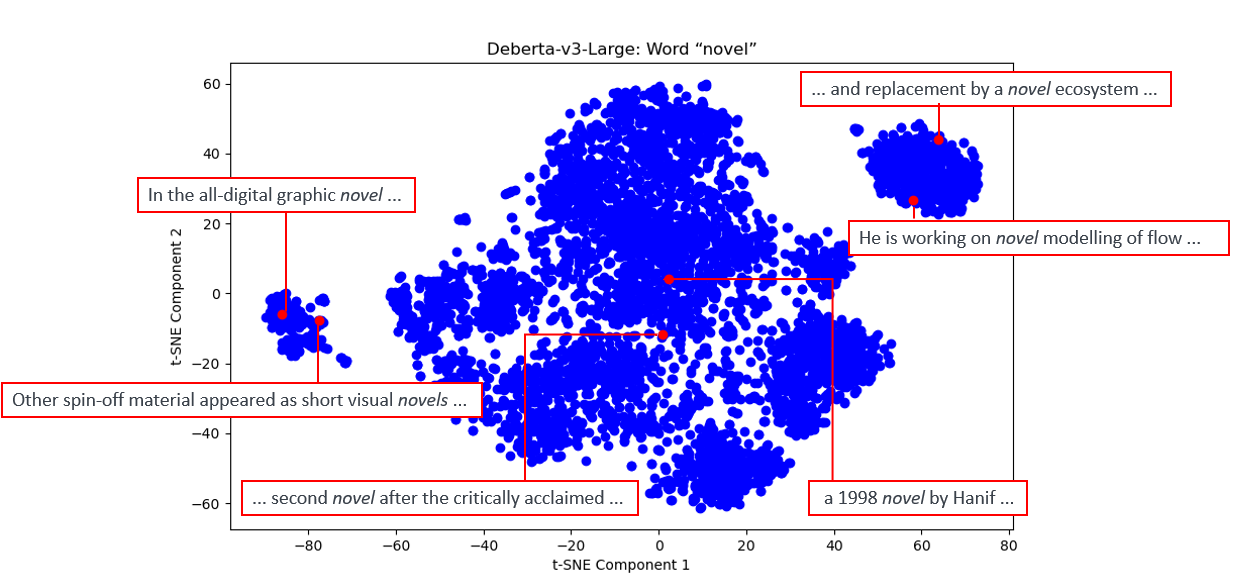} 
\caption{Contextual embeddings of the token \textit{novel} generated by the Deberta-v3-large LLM model~\cite{he2021debertav3} (scatter plot is shown in 2D via t-SNE~\cite{van2008visualizing} on embedding vectors with dimension $d=1024$, which have a total variance of $220.8$). Two main senses can be observed: i) fictitious prose narrative (center group), and ii) new and unusual (top right group). We also observe an auxiliary sense: iii) visual/graphic novel (leftmost group), which can be considered a subclass of sense i).}
\label{fig:novel}
\end{figure*}

In recent years, transformer-based Large Language Models (LLMs) have revolutionized natural language processing by providing powerful capabilities for a wide range of applications \cite{zhao2023survey}. These language models rely on continuous contextual embeddings, which allow for an infinite number of representations for each token. While this approach has proven effective, it contrasts with the way humans perceive language, where each word often carries only a limited number of distinct senses. 
For example, consider the visualization of the different contextual embeddings of the token ``novel'' in Figure~\ref{fig:novel}. We can observe a good amount of variation in the embeddings, 
but there are only two main senses: i) as a noun indicating a fictitious prose narrative, and ii) as an adjective denoting new or original, with some interesting subclasses (e.g., graphic/visual novels). 
Unlike LLMs, which learn languages by processing vast corpora with token-based vocabularies, humans use a different methodology: we first learn the meaning of a new word by remembering a limited number of senses and storing them in our brains; when encountering new text, we determine the appropriate sense of each word based on the context and then integrate these meanings \citep{davis2009complementary,wojcik2013remembering}. 
This observation raises an intriguing question: can we effectively replace the virtually unlimited context-based embeddings generated by large language models for the same token, with a finite set of embeddings more in-tune with human language understanding?

Inspired by this notion, we explore the potential of \emph{multi-sense discrete embeddings}, or just {\em sense embeddings} for short, as a means to compress and store continuous embeddings as multiple distinct vectors for each token. More specifically, our approach collects the embeddings produced from an LLM encoder and feeds them to a clustering algorithm, e.g., K-means clustering \cite{macqueen1967some}, 
which returns the $k$ cluster centers for each token in the vocabulary. These cluster centers serve as the sense embeddings, and are stored in a {\em sense dictionary} to provide compact representations that preserve essential semantic information, and are readily used in subsequent layers of LLMs. Furthermore, we propose a novel sense-based knowledge distillation technique. Knowledge distillation refers to learning a smaller student model that is trained to replicate the behavior of a larger teacher model, thereby facilitating computational time and space savings during inference time. In our sense-based distillation, we use the sense dictionary to replace the last hidden layer in a large (teacher) encoder to yield a much smaller (student) encoder model. 

We conduct empirical experiments on both encoder-based and decoder-based models. 
To show the effectiveness of our multi-sense embeddings we evaluate them as a drop-in replacement in lieu the LLMs' continuous embeddings, and we also study their performance on the word similarity task.
In addition, we evaluate our knowledge distillation framework using the GLUE \citep{wang2019glue} and  MTEB \citep{muennighoff2022mteb} benchmark tasks and demonstrate that significant space savings can be achieved without incurring a major decrease in the observed accuracy. 
Our main contributions can be summarized as follows:

\begin{itemize}
    \item We construct a sense dictionary that bridges the gap between discrete and continuous embeddings. We empirically demonstrate that our sense embeddings can effectively capture semantic information. In addition to enhancing performance in word similarity tests, the proposed sense embeddings retain almost all the information in the LLM drop-in replacement tests.
    
    \item We propose a knowledge distillation method that leverages sense embeddings to guide the student model's learning process. That is, a smaller student model learns to map or choose the correct sense for each token by mimicking the larger teacher model. The proposed knowledge distillation model outperforms the fine-tuned DeBERTa-v3-xsmall~\cite{he2021debertav3} on GLUE benchmarks, and achieves 92\% of the average accuracy while utilizing only 19\% of the GPU memory compared to LLaMA-3-8b-Instruct~\cite{touvron2023llama} on MTEB classification tasks.
\end{itemize}

\section{Background and Related Work}
\noindent {\bf Language Models:} There are mainly two types of language models: discrete embedding-based language models and context embedding-based language models \cite{bommasani2020interpreting,dufter2021static}. 

Discrete embedding-based language models, such as Word2Vec \cite{mikolov2013efficient}, DRG2Vec \cite{shu2020drg2vec}, Dict2Vec \cite{tissier2017dict2vec}, and HG2Vec \cite{wang2022hg2vec} utilize sources like Wikipedia or dictionaries as the input corpus. These models slide a context window along the input text and maximize the similarity of words within each context window. After training, each word is assigned a discrete embedding, allowing users to leverage them directly during inference. But they typically assign one embedding vector per word (or token), and therefore fail to capture context-specific meanings for polysemous words, i.e., those that have different meanings in different contexts (e.g., the word `bank' can refer to a financial institution, a rising ground, an airplane's incline and so on).

Contextual language models \cite{laskar2020contextualized,ganguly2015word} can be grouped into three main types: pure encoder-based models, such as BERT \cite{devlin2018bert} and DeBERTa \cite{he2021debertav3}; pure decoder-based models, such as GPT2 \cite{radford2019language} and Llama \cite{touvron2023llama}; and encoder-decoder-based models, such as T5 \cite{raffel2020exploring}. All of them are based on the attention mechanism \cite{vaswani2017attention}. Encoder-based models transform the input text into a contextualized representation by applying self-attention mechanisms to capture dependencies between tokens across the entire input. Decoder-based models auto-regressively leverage the self-attention mechanism to generate new tokens. Encoder-decoder-based models integrate an encoding mechanism to capture and represent the contextual semantics of the input and then sequentially generate the output sequence through decoders. Nonetheless, even though LLMs excel at understanding the meaning of the entire corpus and demonstrate superior performance on various tasks, they suffer from high space and computational costs for both training and inference.

Our proposed approach aims to leverage the strengths of both discrete embedding-based language models and context embedding-based language models. To achieve this, we gather the output embeddings from LLMs and cluster them into groups to discern the different senses of a token. This approach preserves the semantics, at the same time significantly reduces the time and space requirements during inference.

\smallskip \noindent {\bf Sense Embeddings:}
In contrast to the traditional concept of embeddings, where each word is represented by a single vector, sense embeddings associate multiple vectors per word, where each one of the vectors aims to capture a different meaning \cite{camacho2018word,neelakantan2014efficient}. However, different models utilize different definitions of sense embeddings. MSSG \cite{neelakantan2015efficient} extends the traditional Skip-Gram model to effectively capture multiple senses of a word by clustering the context of word occurrences and associating each cluster with a distinct sense embedding. FastText \cite{athiwaratkun2018probabilistic} introduces multi-sense embedding per word with Gaussian components. 
\citet{amrami-goldberg-2018-word} do word sense induction but use discrete word representations, i.e., as a probability distribution over words, instead of using continuous embeddings, and therefore their sense vectors are not suitable for use in LLM models. AutoExtend~\cite{rothe-schutze-2015-autoextend} focues on earning embeddings for synsets (a group of synonyms) and lexemes (associating spellings with particular meanings), treating both words and synsets as a sum of their lexemes, whereas our focus is on individual token-based sense embeddings, rather than group-based embeddings. \citet{pelevina-etal-2016-making} use ego-network graph-based clustering of word vectors to create sense vectors, which is complimentary to our approach.
LMMS \cite{loureiro2105lmms} generates embeddings per sense key defined by annotated resource, such as WordNet \cite{miller1995wordnet}.  As a result, LMMS can achieve more precise representations for each sense key but lacks support for most datasets consisting of plain text. In contrast, our work develops a general solution to generate multiple embeddings per token without requiring annotation.

\smallskip \noindent{\bf Vector Quantization (VQ):}
VQ is a data compression technique where a large set of vectors is represented by a smaller set of reference vectors, called a codebook (with each vector called a code). \cite{jegou2010product} significantly reduces memory and computation cost for searching by introducing product quantization. Extending this, VQ-VAE \cite{van2017neural} incorporates VQ into deep generative models through Variational Autoencoders, showcasing its potential for learning discrete latent representations. Transformer-VQ \cite{lingle2023transformer} proposes a linear time attention model for decoders with using vector-quantized keys and a novel caching mechanism within the attention process. Our work is closely related to VQ, but instead of a global codebook, our sense dictionary can be considered as using a fine-grained token-based codebook to elicit and capture a finite set of meanings or senses for each token. Incidentally, our work is complementary and compatible with the precision-based quantization approaches that use low-bit representation to optimize learning cost~\cite{Dettmers2022LLMint88M, chen2024efficientqat}.

\smallskip \noindent{\bf Knowledge Distillation: }
Knowledge distillation is a key model compression technique that trains a smaller student model to replicate the outputs of a larger teacher model. The primary advantage of this approach is its ability to significantly reduce model size without substantially sacrificing performance. Various methods have been developed to enhance this training process. The first approach focuses on replicating the distribution of the output layer, such as DistilBERT \cite{Sanh2019} and LIGHTPAFF \cite{song2020lightpaff}. The second category leverages the benefits of mimicking the hidden states. TinyBERT \cite{jiao2020tinybert} and PKD \cite{sun2019patient} leverage the information in the teacher's hidden layers, encouraging the student model to learn through a multi-step distillation process. MINILMv2 \cite{wang2021minilmv2} introduces an additional strategy of mimicking the attention layers. 
Recently, with the emergence of extremely large language models containing hundreds of billions of parameters, methods like Alpaca \cite{taori2023stanford} and Vicuna \cite{chiang2023vicuna} have been developed to mimic the outputs generated by a larger LLM via instruction or conversation-based fine-tuning. We propose a novel approach to knowledge distillation, where the much smaller student model learns to choose the correct sense for each token based on the guidance from a much larger teacher model. This can significantly cut down on the space cost with little loss in performance.


\section{Multi-sense Embeddings}

We first tackle the issue of whether we can extract meaningful senses for the tokens and use a discrete set of sense embeddings per token, rather than having an unlimited number of representations for the same token. We next tackle the issue of using our multi-sense embeddings for a novel sense-based knowledge distillation approach whereby a smaller student model learns to mimic the sense output of the larger teacher (base) LLM, thereby saving both space and time during inference.

As a motivating example, we 
consider the DeBERTa-v3-large LLM~\cite{he2021debertav3}, which is an encoder-based model with hidden dimensionality $d=1024$, and it improves over both BERT~\cite{devlin2018bert} and RoBERTa~\cite{liu2019robertarobustlyoptimizedbert} models using disentangled attention, enhanced mask decoder, and gradient disentangled embedding sharing. 
We randomly selected 5,000 occurrences of the token \textit{novel}, from the English Wikipedia dump~\cite{wikipedia_dump_20240320}; it has two primary meanings: \textit{new and unusual} and \textit{a fictitious prose narrative}. 
Figure~\ref{fig:novel} shows the distribution of contextual embeddings for \textit{novel} from the last hidden layer of DeBERTa-v3-large \cite{he2021debertav3}, where we use t-SNE \cite{van2008visualizing} to visualize the embeddings in 2D. 
We see a considerable spread of these embeddings in the hidden latent space. Nevertheless, tokens with similar meanings naturally form clusters, rather than being randomly distributed. 
The larger cluster in the center corresponds to the meaning \textit{a fictitious prose narrative}, while the smaller cluster on the top right represents \textit{new and unusual}. 
We also observe subgroups depending on the context, e.g., the smaller cluster on the left represents phrases such as \textit{graphic novel} and \textit{visual novel} is a subgroup of the first meaning.
We conclude that clustering can capture semantic information based on context or usage, and the distinct centers can represent different senses of each token.

\subsection{Sense Dictionary}

\begin{figure}[t]
\centering
\includegraphics[width=0.5\textwidth]{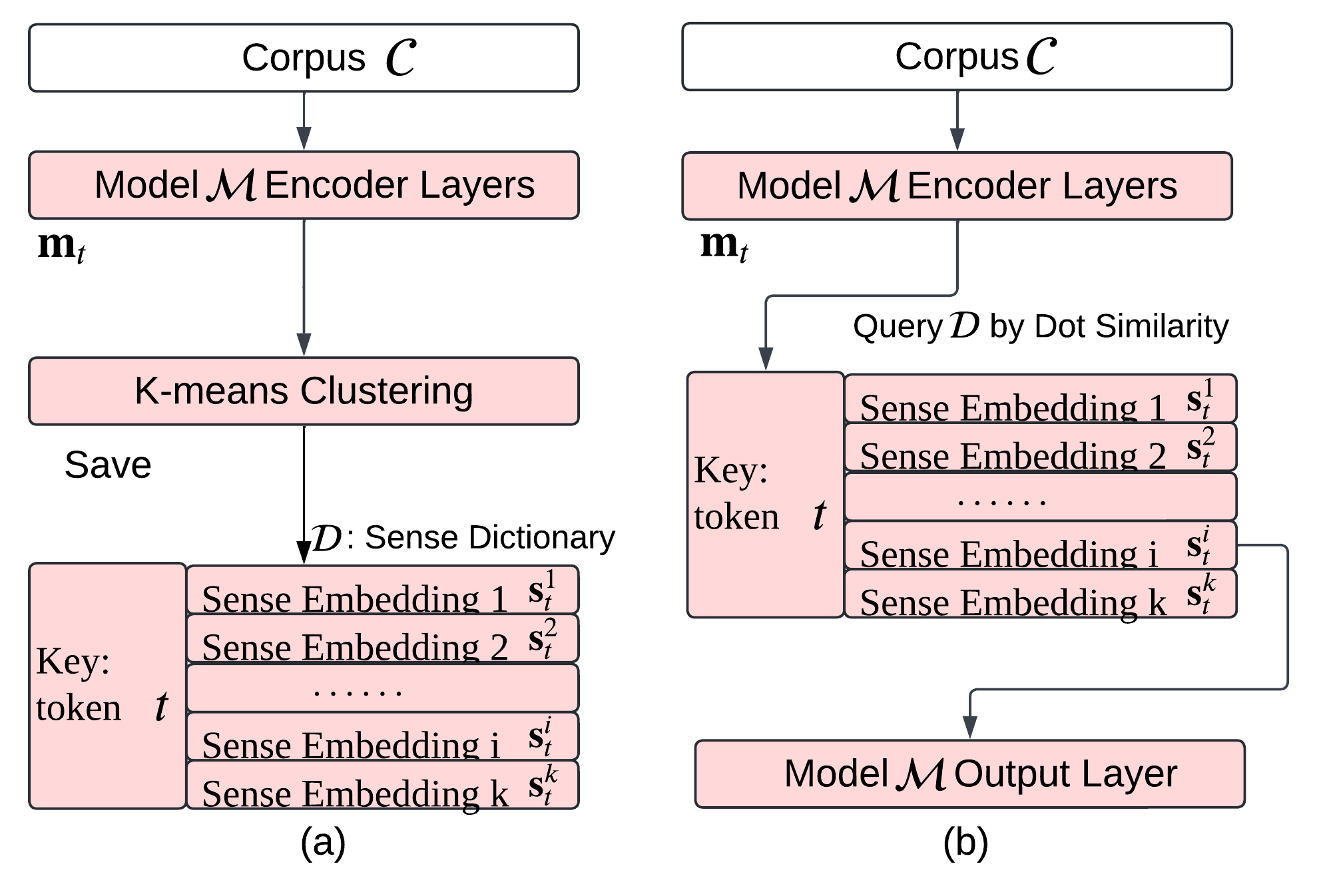} %
\caption{(a) Exploiting cluster centers for sense embedding dictionary. (b) LLM replacement test.}
\label{fig:cluster}
\end{figure}

Let $\mathcal{C}$ denote the input corpus, which is fed into the encoder of a pretrained large language model $\mathcal{M}$. For each token $t$ in the model's vocabulary $\mathcal{V}$, let $E_t$ denote the set of all of its embeddings from the last hidden layer of the model, given as $E_t = \{ \mathbf{m}_t^i \}_{i=1}^{n_t}$, where $n_t$ is the number of occurrences of token $t$ in the corpus, and $\mathbf{m}_t^i$ is the contextual embedding for the $i$-th occurrence.  Next, we use K-means~\cite{macqueen1967some} to cluster the set of embeddings $E_t$ for each token $t$. The resulting $k$ cluster centroids (in latent space) represent distinct sense embeddings, denoted as $\mathbf{s}_t = \{\mathbf{s}_{t}^{(i)}\}_{i=1}^{k}$, where each centroid $\mathbf{s}_{t}^{(i)}$ corresponds to a unique discrete sense of token $t$.
Finally, we construct a multi-sense embedding dictionary $\mathcal{D}$ for the model $\mathcal{M}$. This dictionary  $\mathcal{D}$ stores for each token $t$, its corresponding set of multi-sense embeddings $\mathbf{s}_t$. 
Thus, $\mathcal{D} = \{\mathbf{s}_t \}$ for $t \in \mathcal{V}$.
Figure~\ref{fig:cluster}(a) illustrates the sense dictionary construction step from the LLM. The discussion above was in the context of encoder-based LLMs. To handle decoder-based models, we leverage LLM2Vec~\cite{behnamghader2024llm2vec} to transform decoders into encoders and follow the same process.

\begin{figure*}[th]
\centering
\includegraphics[width=0.8\textwidth, height=3in]{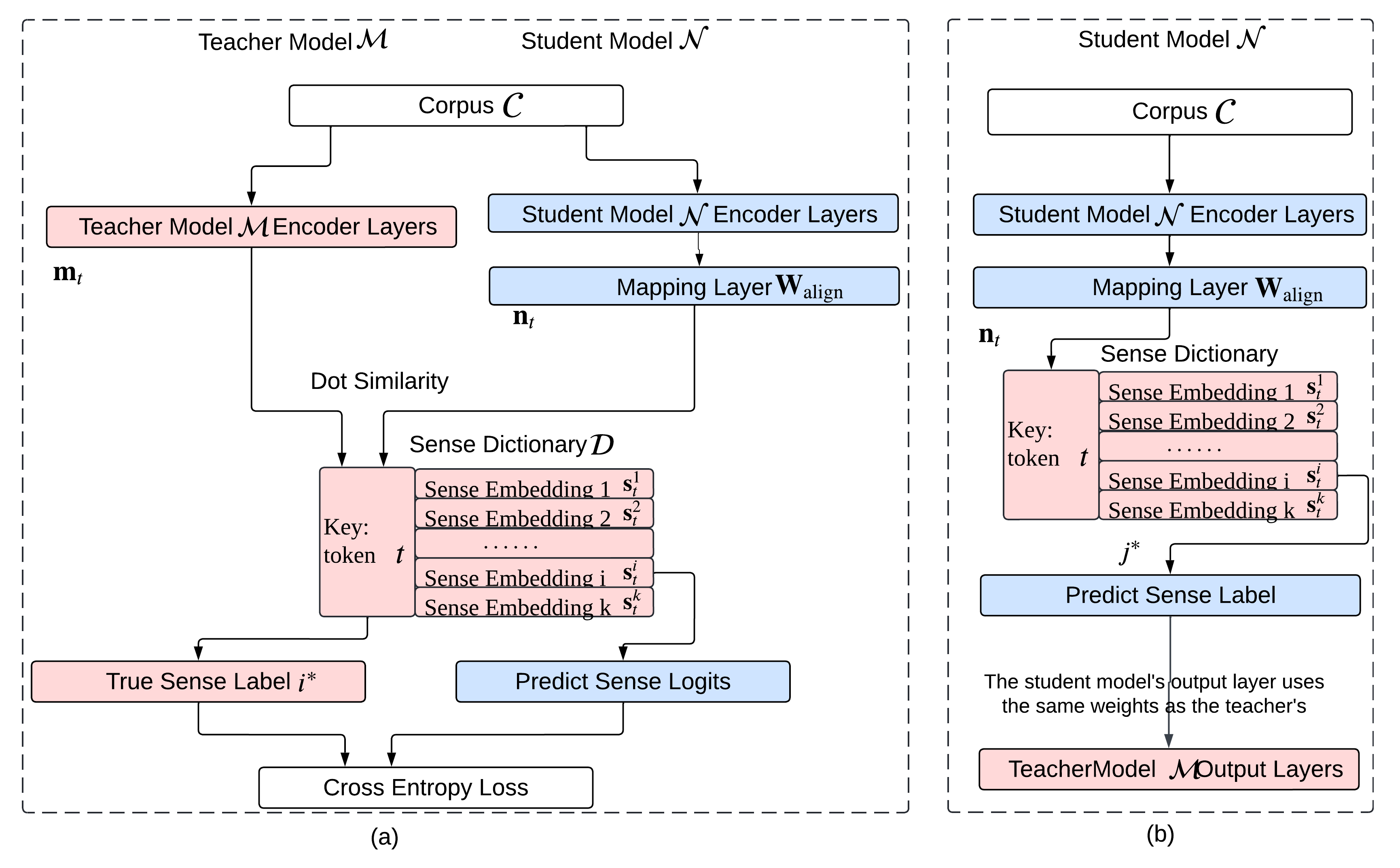} 
\caption{Knowledge distillation process. (a) Training process. (b) Evaluation process.}
\label{fig:knowledge_distillation}
\end{figure*}


\subsection{Drop-in LLM Replacement}
\label{sec:dropin}
Having obtained the sense dictionary $\mathcal{D}$ from the {\em training corpus} $\mathcal{C}$, we now explore how we can replace the continuous contextual embeddings from the LLM with our sense embeddings.
We begin by passing the {\em test corpus} $\mathcal{C}_T$ through the encoder model $\mathcal{M}$ to obtain the continuous embedding $\{\mathbf{m}_t\}$ for each occurrence of the token $t$.
Next, we replace the continuous embedding $\mathbf{m}_t$ with the sense embedding $\mathbf{s}_t^{(i^*)}$ that maximizes the dot product similarity:
$i^* = \arg\max_{i} \big\{ \langle\mathbf{m}_t, \mathbf{s}_t^{(i)}\rangle \big\}$.

This selected sense $\mathbf{s}_t^{(i^*)}$ serves as the discrete embedding for that token occurrence. Thus, the sequence of discrete embeddings 
$\{\mathbf{s}_{t_j}^{(i^*)} \}$ for each token $t_j$ in the context block replaces the sequence of continuous embeddings $\{ \mathbf{m}_{t_j} \}$ from the last hidden layer. These discrete sense embeddings are then propagated through the subsequent layers of the model according to its architecture. Finally, we collect the modified model's output and compare it to the output of the base LLM to assess the effectiveness of the approach. Figure \ref{fig:cluster}(b) illustrates the drop-in LLM replacement pipeline.
In our empirical studies in Sec.~\ref{sec:exp}, we demonstrate that sense embeddings can effectively serve as a drop-in replacement for contextual embeddings with little to no loss in downstream performance.

\subsection{Sense-based Knowledge Distillation}
 \label{sec:kd}
 We now propose our novel approach for knowledge distillation based on our sense dictionary. In knowledge distillation, the task is to learn a smaller student model that is trained to replicate the behavior of a larger teacher model. During inference, only the student model is employed, which facilitates both reduced memory and compute costs. The multi-sense token embeddings in our sense dictionary act as a guide for the student model's learning process.
 To our knowledge, this is the first approach that proposes a sense-based knowledge distillation framework. In essence, we convert the distillation task into a classification task that trains the student model to select the same candidate from the sense dictionary as identified by the teacher model. Subsequently, the student model can be employed for any downstream task.

\smallskip \noindent {\bf Student Model Training:}
The knowledge distillation step for student model training is illustrated in Fig.~\ref{fig:knowledge_distillation}(a).
Both the teacher model $\mathcal{M}$ and the student model $\mathcal{N}$ process the input training corpus $\mathcal{C}$ through their respective encoders, yielding continuous embeddings,  $\{\mathbf{m}_t\}$ and 
$\{\mathbf{n}_t\}$. 
Note that the student model can employ smaller hidden layer dimensionality, and for this reason, the student embeddings are passed through the mapping layer $\mathbf{W}_{\text{align}}$ to 
align their dimensionality with the teacher model (if needed).
Next, the teacher model employs the 
sense dictionary $\mathcal{D}$ to retrieve the corresponding set of multi-sense embeddings $\mathbf{s}_t$, and computes the dot product similarity between the continuous embedding $\mathbf{m}_t$ and each sense embedding for token $t$, selecting the sense embedding with the highest similarity:
$i^* = \arg\max_{i} \big\{
    \langle \mathbf{m}_t, \mathbf{s}_t^{(i)} \rangle \big\}$.
For knowledge distillation, the sense $\mathbf{s}_t^{(i^*)}$ selected by the teacher model is assumed to be correct, and this sense's index $i^*$ is used as the {\em ground truth} sense label for that token. 

To train the student model, we employ Cross Entropy Loss, maximizing the probability that the student model selects the same sense embedding as the teacher model by minimizing the difference between the teacher's sense embedding $\mathbf{s}_t^{(i^*)}$ and the student's continuous output from encoder $\mathbf{n}_t$. The loss is given as:
$\mathcal{L}_{\text{CE}} = -\sum_{t} \log P\big(\mathbf{s}_t^{(i^*)} \mid \mathbf{n}_t\big)$, 
where $(i^*)$ is the true sense label from the teacher model for $t$-th token, and $P\big(\mathbf{s}_t^{(i^*)} \mid \mathbf{n}_t\big)$ is the probability of the student model's output matching the teacher's selection, obtained via the softmax function:
$P\left(\mathbf{s}_t^{(i^*)} \mid \mathbf{n}_t\right) = \frac{\exp\left(\langle \mathbf{n}_t, \mathbf{s}_t^{(i^*)} \rangle\right)}{\sum_{j=1}^k\exp\left(\langle \mathbf{n}_t, \mathbf{s}_t^{(j)} \rangle\right)}$.

\medskip \noindent {\bf Student Model Inference:}
During evaluation or inference process, the teacher model $\mathcal{M}$ is no longer involved, as illustrated in Fig.~\ref{fig:knowledge_distillation}(b).
Given the test corpus $\mathcal{C}_T$, 
the student model $\mathcal{N}$ processes the input through its encoder and the mapping layer $\mathbf{W}_{\text{align}}$, producing output embeddings $\{\mathbf{n}_t\}$. For 
each token $t$, the student model selects the sense embedding 
$\mathbf{s}_t^{(j^*)}$ from the dictionary $\mathcal{D}$ by maximizing the dot product similarity given as:
$j^* = \arg\max_{j} \big\{ 
    \langle \mathbf{n}_t, \mathbf{s}_t^{(j)} \rangle \big\}$.
The selected sense embeddings $\lbrace\mathbf{s}_t^{(j^*)}\rbrace$ over all tokens in the context window replace the last hidden layer's continuous embeddings. 
Subsequently, these embeddings are passed through the output layer, which shares the same architecture and weights as the teacher’s model $\mathcal{M}$. Finally, the results are collected and evaluated using the same metrics as those for the teacher model $\mathcal{M}$. 

\section{Empirical Evaluation}
\label{sec:exp}

For our experiments we employ up to 6 NVIDIA V100 GPUs (32GB RAM each) with 20-core IBM Power 9 processors (512GB RAM). Experimental details on the various benchmarks and models are given in the Appendix. Our code is available at \url{https://github.com/Qitong-Wang/SenseDictionary}.

\subsection{Drop-in LLM Replacement Effectiveness}

We first test the effectiveness of our sense dictionary by utilizing it for drop-in replacement for the last hidden layer's output by mapping the continuous embeddings to the discrete sense embeddings, as outlined in Sec.~\ref{sec:dropin}.

\begin{table}[ht]
\centering
    \resizebox{0.3\textwidth}{!}{%
    \begin{tabular}{|l|c|c|}
    \hline
    \textbf{GLUE Task} & \textbf{DeBERTa-v3} & \textbf{Drop-in} \\ \hline
    CoLA               & 75.72                    & 73.13                         \\ \hline
    MNLI               & 91.64                    & 91.10                          \\ \hline
    MRPC               & 92.85                    & 91.98                         \\ \hline
    QNLI               & 95.51                    & 94.51                         \\ \hline
    QQP                & 92.00                       & 92.06                         \\ \hline
    RTE                & 90.61                    & 88.81                         \\ \hline
    SST-2              & 96.21                    & 96.21                        \\ \hline
    STS-B              & 92.93                    & 90.48                         \\ \hline
    \end{tabular}
    }
\caption{LLM Replacement Test for DeBERTa-v3-large.}
\label{tab:llm_replacement}
    \end{table}

We select the DeBERTa-v3-large model \cite{he2021debertav3} as the representative encoder model due to its state-of-the-art performance on GLUE benchmark classification tasks \cite{wang2018glue}. We construct the sense dictionary using the combination of training datasets of GLUE and apply K-means clustering \cite{macqueen1967some} with $k=15$ to extract the sense embeddings dictionary $\mathcal{D}$.  
During inference, we utilize the development datasets from the benchmark to evaluate the original (base) LLM versus the replaced model, where the continuous output embeddings have been replaced by the best matching senses from $\mathcal{D}$.
From the results in Table~\ref{tab:llm_replacement}, we can clearly see that replacing continuous embeddings with discrete sense embeddings achieves nearly identical performance across most datasets.

\begin{table}[ht]
\centering
\resizebox{0.45\textwidth}{!}{%
\begin{tabular}{|l|c|c|}
\hline
\textbf{Task} & \textbf{Llama3-8B-Instruct} & \textbf{Drop-in} \\
\hline
Arxiv                & 74.09 & {\em 77.57} \\ \hline
Banking77            & 88.01 & 79.04 \\ \hline
DBpedia              & 92.29 & 90.05 \\ \hline
Emotion              & 51.29 & 49.91 \\ \hline
FrenkEn              & 69.25 & 65.65 \\ \hline
Imdb                 & 83.23 & {\em 88.33} \\ \hline
LegalBench           & 90.16 & 86.89 \\ \hline
News                 & 80.11 & {\em 84.49} \\ \hline
Patent               & 40.31 & 38.43 \\ \hline
PoemSentiment        & 53.08 & 47.79 \\ \hline
ToxicChat            & 82.87 & 75.65 \\ \hline
ToxicConversations   & 67.78 & 66.38 \\ \hline
TweetSentimentExtraction & 61.84 & 60.00 \\ \hline
TweetTopicSingle     & 70.47 & 71.72 \\ \hline
YelpReviewFull       & 53.94 & 54.68 \\ \hline\hline
\textbf{Average}                   & 70.58 & 69.10 \\ \hline
\end{tabular}
}
\caption{LLM Replacement Test for  Llama3-8B-Instruct on MTEB Classification Tasks.}

\label{tab:llm_replacement2}
\end{table}

\begin{table*}[ht]
\centering
\resizebox{0.9\textwidth}{!}{%
\begin{tabular}{|l|cc|ccc|}
\hline
\textbf{Dataset} & \textbf{DeBERTa-v3-large} & \textbf{DeBERTa-v3-xsmall} & \textbf{SKD} & \textbf{SKD w/o}  & \textbf{DistilDeBERTa} \\
 &  &  &  & \textbf{SenseDict} &  \\
\hline
CoLA & 75.72 & 64.16+/-0.87 & \textbf{65.50+/-0.89} & 64.13+/-0.48 & 62.48+/-0.38 \\
\hline
MNLI & 91.64 & 88.29+/-0.07 & \textbf{87.15+/-0.18} & 85.80+/-0.17 & 86.03+/-0.24 \\
\hline
MRPC & 92.85 & 90.19+/-0.90 & 90.45+/-0.26 & 89.10+/-0.41 & \textbf{91.06+/-0.41}\\
\hline
QNLI & 95.51 & 92.12+/-0.54 & \textbf{91.63+/-0.31} & 91.60+/-0.29 & 91.29+/-0.39 \\
\hline
QQP & 92.00 & 89.08+/-0.06 & \textbf{90.72+/-0.12} & 90.64+/-0.30 & 90.12+/-0.32 \\
\hline
RTE & 90.61 & 74.36+/-1.06 & \textbf{77.13+/-0.44} & 73.70+/-0.95 & 72.56+/-0.88 \\
\hline
SST-2 & 96.21 & 93.43+/-0.30 & \textbf{91.68+/-0.29} & 89.68+/-0.80 & 89.34+/-0.57 \\
\hline
STS-B & 92.93 & 89.43+/-0.26 & 87.41+/-0.21 & \textbf{88.70+/-0.39} & 88.40+/-0.16 \\
\hline

{\bf \# of Parameters}   & 300M   & 22M      & 22M     & 22M  & 22M  \\ \hline
{\bf GPU Memory} & 600MB & 44MB & 44MB & 44MB  & 44MB\\\hline
\end{tabular}
}
\caption{Knowledge distillation on DeBERTa-V3-large: Comparison of our sense-based knowledge distillation student model, denoted SKD, with the original model, the Distil-DeBERTa distilled model, and our SKD model without sense dictionary (SKD w/o SenseDict).}
\label{tab:glue_tasks}
\end{table*}

Next, we use the Llama3-8B-Instruct model \cite{touvron2023llama} as a representative of decoder-based LLM. Decoder models use auto-regressive output generation, where only the previous tokens are used to construct the continuous embeddings at the current token in the context. This is not fully compatible with the encoder models that look at the entire context, and therefore, to effectively utilize the same sense dictionary, which is based on the full context, we first utilize LLM2Vec~\cite{behnamghader2024llm2vec} to convert the decoder into an encoder model.
We evaluate the original Llama3-8B-Instruct model and its drop-in replacement version on the MTEB classification tasks \cite{muennighoff2022mteb}. We build the sense dictionary from the MTEB training datasets. For clustering, some very generic tokens may require a larger number of senses. We propose a dynamic clustering method to determine the optimal number of clusters for each token. Specifically, we utilize the Markov Clustering (MCL) \cite{van2008graph} approach to estimate the initial number of clusters. However, MCL tends to produce an excessively high number of clusters, so we introduce a scaling coefficient to adjust the cluster count derived from MCL, followed by the application of K-means clustering. As a result, each token is assigned a different $k$, ensuring a more flexible and adaptive clustering strategy. Further implementation details can be found in the Appendix.
We can observe from Table~\ref{tab:llm_replacement2} that there is very little performance drop when using the discrete sense embeddings in lieu of the continuous embeddings. Interestingly, for {\em Arxiv}, {\em Imdb}, and {\em News} the discrete senses even lead to better performance.
These results confirm that using a sense dictionary to replace continuous embeddings maintains performance while effectively capturing the most useful contextual information.

\subsection{Knowledge Distillation Effectiveness}

We now examine the effectiveness of our novel sense-based knowledge distillation approach, where we train the student model to mimic the sense choice from the teacher model as described in Sec.~\ref{sec:kd}. We denote our student model as SKD (for 
{\bf S}ense-based {\bf K}nowledge {\bf D}istillation). We selected the GLUE benchmark \cite{wang2018glue} to compare with DeBERTa-v3-large\cite{he2021debertav3}, and the MTEB classification benchmark \cite{muennighoff2022mteb} for LLama3-8B-Instruct \cite{touvron2023llama} because their respective papers and codebases use these benchmarks. Therefore, we can leverage pre-trained teachers without retraining, and establish a fair comparison.

\subsubsection{\bf Encoder LLMs}
We compare our approach with several models to ensure a fair and consistent evaluation: the fine-tuned DeBERTa-v3-large \cite{he2021debertav3} as the teacher model, the smaller DeBERTa-v3-xsmall as the architectural baseline, our proposed sense-based student model (SKD), a variant SKD w/o SenseDict—which excludes the sense embedding dictionary and instead relies solely on CrossEntropyLoss to align student and teacher representations, and a distilled version of DeBERTa following the DistilBERT methodology (Distil-DeBERTa) \cite{sanh2019distilbert}. As previously mentioned, we construct the sense dictionary $\mathcal{D}$ using the combined training sets from all GLUE classification tasks \cite{wang2018glue}. Since DeBERTa is fine-tuned individually for each task, we follow the same procedure by training a separate SKD model for each subtask using only its respective dataset. To maintain consistency, all student models (SKD, SKD w/o SenseDict, and Distil-DeBERTa) adapts the same architecture of DeBERTa-v3-xsmall, with 6 layers and a hidden dimension of 384 (Deberta-v3-large has 24 layers and a hidden dimension of 1024). We report the mean and standard deviation over three independent runs, enabling a robust and fair comparison across all settings.

Table~\ref{tab:glue_tasks} presents the results of knowledge distillation on the GLUE benchmark, comparing our SKD model with SKD w/o SenseDict, DistilDeBERTa, and DeBERTa-v3-xsmall. Overall, SKD achieves competitive or superior performance across multiple tasks, demonstrating the effectiveness of incorporating multi-sense embeddings in knowledge distillation. Notably, SKD exhibits a significant improvement on the RTE dataset, surpassing SKD without sense embedding dictionary by 3.43\% and even outperforming DistilDeBERTa by a considerable margin. 
This indicate that multi-sense embeddings enhance the model’s ability to generalize on tasks with limited training data, where traditional distillation methods struggle due to insufficient task-specific knowledge.

\paragraph{Case Study:}
To analyze the results of our sense distillation we examine cases where SKD classified correctly or incorrectly compared to the baselines. We look at RTE, which requires a binary classification to determine whether a given sentence (hypothesis) is entailed by another sentence (premise). Consider the instance, where for the premise \textit{It has been observed that in those countries of the world where capital punishment is still in operation, the crime rate, especially murder, is distinctively low in comparison to countries where capital punishment has been discarded}, the hypothesis is \textit{ Capital punishment is a deterrent to crime.} The correct label is \textit{ entailment}. Both DeBERTa-V3-large and our knowledge distillation model SKD correctly classify this instance, whereas DeBERTa-V3-xsmall and SKD w/o SenseDict fail to do so. This case is challenging because the hypothesis does not explicitly state the relationship between capital punishment and crime; instead, it presents a comparison between two countries with different policies. DeBERTa-V3-xsmall and SKD w/o SenseDict struggles to draw a conclusion based on this comparison.

As another example, consider the premise \textit{... They killed a teacher and 12 students and wounded 23 others before committing suicide. The massacre shocked the country ...'}, with the hypothesis \textit{13 persons were killed by two students in 1999.} The correct label is \textit{entailment}. In this example, DeBERTa-V3-large does the correct prediction, whereas SKD, SKD w/o SenseDict, and DeBERTa-V3-xsmall get it wrong. As we can see, this entailment case requires adding the number of people, where the smaller models don't do as well.

As such, logical reasoning tasks, as in RTE, benefit from a larger and more capable model like DeBERTa-v3-large. In contrast, DeBERTa-v3-xsmall, due to its significantly reduced size, struggles more to capture and generalize logical relationships effectively. Our SKD model, which transfers knowledge from DeBERTa-v3-large, can successfully solve some logical entailments, but not the more intricate ones, given its size.

\begin{table*}[ht]
\centering
\resizebox{0.85\textwidth}{!}{%
\begin{tabular}{|l|c|ccc|cc|}
\hline

{\bf Dataset}                   & {\bf LLM2Vec}            & {\bf SKD}            & {\bf SKD w/o}        & {\bf SKD-Sep} & {\bf Train File} & {\bf \# of Train} \\
                          & {\bf Llama3-8B-Instruct} &                & {\bf SenseDict}      &         & {\bf Size (MB)}  & {\bf Sentences} \\
\hline
Arxiv                     & 74.09              & \textbf{76.55} & 71.86          & 76.51   & 111.00     & 29268 \\
\hline
Patent                    & 40.31              & \textbf{36.76} & 35.89          & 36.67   & 99.00      & 25720 \\
\hline
News                      & 80.11              & \textbf{83.62} & 80.27          & 83.70   & 78.00      & 120320 \\
\hline
Imdb                      & 83.23              & \textbf{83.63} & 77.85          & 83.34   & 62.00      & 25176 \\
\hline
ToxicConversations        & 67.78              & \textbf{74.04} & 68.48          & 67.49   & 43.00      & 50320 \\
\hline
YelpReviewFull            & 53.94              & \textbf{50.01} & 45.00          & 52.37   & 13.00      & 8448 \\
\hline
Emotion                   & 51.29              & 49.57          & \textbf{50.56} & 50.91   & 11.00      & 16960 \\
\hline
Banking77                 & 88.01              & 74.58          & \textbf{78.61} & 78.27   & 6.20       & 16163 \\
\hline
FrenkEn                   & 69.25              & 63.66          & \textbf{64.16} & 64.21   & 4.20       & 8564 \\
\hline
DBpedia                   & 92.29              & \textbf{87.48} & 87.28          & 90.15   & 3.30       & 3168 \\
\hline
TweetSentimentExtraction  & 61.84              & 58.60          & \textbf{60.01} & 60.30   & 3.10       & 4494 \\
\hline
ToxicChat                 & 82.87              & 74.04          & \textbf{75.18} & 77.49   & 3.10       & 2961 \\
\hline
TweetTopicSingle          & 70.47              & 58.60          & \textbf{60.99} & 70.63   & 2.40       & 1996 \\
\hline
LegalBench                & 90.16              & \textbf{64.34} & 50.41          & 84.42*  & 1.80       & 1058 \\
\hline
PoemSentiment             & 53.08              & 44.42          & \textbf{48.17} & 48.27   & 1.70       & 1212 \\
\hline
\hline
\textbf{Average}          & 70.58              & \textbf{65.33} & 63.65          & 68.31   &            & \\ \hline
\hline
{\bf \# of Parameters}    & 7.5B               & 1.4B           & 1.4B           & 1.4B    &            & \\ \hline
\textbf{GPU Memory}       & 15GB               & 2.7GB          & 2.7GB          & 2.7GB   &            & \\
\hline
\end{tabular}
}
\caption{Performance Comparison between Llama3-8B-Instruct and our SKD Student Model and its variants SKD w/o SenseDict and SKD-Sep. For LegalBench with SKD-Sep, we combine the LegalBench dataset with a randomly sampled subset from other datasets.}
\label{tab:comparison_filtered}
\end{table*}

\subsubsection{\bf Decoder LLMs}
We select Llama3-8B-Instruct \cite{touvron2023llama} as a representative of decoder models and apply the LLM2Vec wrapper \cite{behnamghader2024llm2vec} to transform it into an encoder. We choose classification tasks from the MTEB benchmark \cite{muennighoff2022mteb}, and we iterate through the entire training dataset to build the sense dictionary. For our SKD student model, we simplify the architecture by using only 4 layers of the Llama3-8B-Instruct model, as opposed to the original 32 layers. This reduction results in a significant decrease in memory usage: the student model requires only 2.7GB of GPU memory to store the model weights, whereas the full Llama3 model needs 15GB GPU memory due to its larger number of layers. Note that whereas the full sense dictionary requires 8.4GB CPU memory, much smaller GPU memory is needed for the subset of the dictionary for tokens in the current context window (e.g., under 670MB for the larger Llama3 model; see Appendix for details). Since Llama3 is a general-purpose language model that utilizes a single model for all downstream tasks, we combine all training datasets from the MTEB classification tasks and train a single student model for classification tasks, denoted as SKD in Table \ref{tab:comparison_filtered}. Meanwhile, SKD w/o SenseDict represents the knowledge distillation model that does not incorporate sense embedding dictionary and SKD-Sep refers to the setting where we train a separate student model for each individual classification task using only that task’s dataset.


Our proposed SKD student model achieves competitive performance across a variety of classification tasks while significantly reducing computational costs. On average, SKD attains 92\% of the accuracy of the full LLM2Vec Llama3-8B-Instruct model, despite utilizing only 19\% of the GPU memory. This demonstrates that SKD maintains high accuracy while being much more memory-efficient, making it suitable for deployment in resource-constrained environments. A closer analysis reveals that SKD consistently outperforms SKD w/o SenseDict, particularly when the training dataset is large. For instance, on the Arxiv dataset, SKD improves performance by 4.69\%, while on the IMDB dataset, it achieves a 5.78\% gain over SKD w/o SenseDict. This suggests that incorporating the sense embedding dictionary enhances knowledge distillation, leading to better generalization on larger datasets. Additionally, while SKD-Sep exhibits advantages in certain datasets, especially when the training dataset is small, its effectiveness diminishes as dataset size increases. This is likely due to the limitation of task-specific knowledge available for each independent student model. In contrast, SKD excels on large datasets by effectively capturing and transferring broader linguistic knowledge from the teacher model, making it more robust across diverse inputs. In conclusion, our SKD approach offers a well-balanced tradeoff between efficiency and accuracy, demonstrating its capability to generalize effectively across multiple classification tasks. This makes SKD a scalable and practical solution for real-world NLP applications where deploying larger models is computationally expensive.

%


\section{Conclusion and Future Work}

In this paper, we make two main contributions. The first is the notion of a sense dictionary that encapsulates multi-sense embeddings that capture a finite number of meanings, as opposed to the infinite number of continuous contextual embeddings in LLM models. 
We show that our sense embeddings can serve as drop-in replacement for the embeddings in both encoder- and decoder-based LLMs, and retain the virtually the same performance. We also propose a novel sense-based knowledge distillation approach, where the significantly smaller student model learns the correct sense based on the teacher LLM, resulting in a significant reduction in space and inference time, but with better performance compared to other baseline distilled models. There are 
still areas for improvement, such as developing an end-to-end approach to learn the sense embeddings. Our sense embedding approach offers the intriguing possibility of learning from a small but diverse corpus that encapsulates most of the senses of each token, rather than relying on trillions of tokens as done by the current generation of LLMs; we plan to explore this in the future.

\section{Limitations}



Our approach has certain limitations that we acknowledge and plan to address in future studies:

\begin{enumerate}
    \item \textbf{Scope of Downstream Tasks:} While our proposed knowledge distillation method is designed to support a wide range of downstream tasks, we only demonstrate its effectiveness on classification tasks in this paper. Our distillation approach focuses on enabling the student model to learn the correct mapping of sense embeddings, which is inherently independent of specific downstream tasks. This flexibility suggests that our method has the potential to perform well across various types of tasks, such as regression, sequence labeling, and generation. However, due to resource and time constraints, we have limited our evaluation to classification tasks. In future work, we plan to extend our experiments to include diverse downstream tasks to further validate the generalizability of our method.

    \item \textbf{Dependency on the LLM2Vec Wrapper:} Our current implementation relies on the LLM2Vec wrapper for the decoder models to convert them into encoder models. While LLM2Vec provides competitive results and even outperforms baseline decoders on some tasks, we plan to study how once can use the causal embeddings from decoders directly to learn decoder sense embeddings, and compare with our current approach.
\end{enumerate}

\bibliography{main}
\appendix
\section*{Appendix}

\section{Open Source Code}
Here is the  link to our code: 
\url{https://github.com/Qitong-Wang/SenseDictionary}.
Our code uses NumPy~\cite{harris2020array}, PyTorch~\cite{paszke2019pytorch}, scikit-learn~\cite{scikit-learn} and MCL~\cite{van2008graph} libraries to process the data and train the model, and Matplotlib~\cite{hunter2007matplotlib} to generate the plots. 

The potential risks of our project are the same as underlined LLMs, such as DeBERTa and LLaMA. Our work focuses on compressing these large models and developing computationally efficient methods; therefore, we do not introduce any additional risks beyond those already present in the underlying models.

\section{Models and Benchmark Datasets}

We select K-means~\cite{macqueen1967some, lloyd1982least, hartigan1975clustering} from scikit-learn~\cite{scikit-learn} and MCL~\cite{van2008graph} for clustering the contextual embeddings. Both are open source.

We select the GLUE benchmark~\cite{wang2018glue}, MTEB classification tasks~\cite{muennighoff2022mteb}, and WikiText~\cite{merity2016pointer} as datasets for knowledge distillation. WikiText is sourced from HuggingFace (\url{https://huggingface.co/}). All of these datasets are open source.

We choose the DeBERTa-v3-large model~\cite{he2021debertav3} as a representative encoder model, the Llama3-8B-Instruct model~\cite{touvron2023llama} as a representative decoder model, and use LLM2Vec~\cite{behnamghader2024llm2vec} to convert it into an encoder.  For the knowledge distillation evaluation, we compare with Distil-DeBERTa model as a baseline knowledge distillation model, which is trained using the DistilBERT~\cite{sanh2019distilbert} methodology. In addition we use DeBERTa-v3-xsmall, since our student model has the same number of parameters.  All of these models, except Llama, are open source. We obtained a license from the LLaMA team and did not modify the source code of Llama.

\section{Sense Dictionary: Parameters}

Table~\ref{tab:glue_wikitext_stats} and Table~\ref{tab:llama_mteb_stats} present the parameters and statistics for the sense dictionary for the DeBERTa-v3-large and Llama3-8B-Instruct model. It is important to note that some tokens, such as foreign language characters or emojis, are not included in our sense embedding process, and some tokens may not occur in the corresponding GLUE and MTEB benchmark training datasets. Consequently, the number of valid tokens is smaller than the vocabulary size of the tokenizer, which is 128K for both models.
We can see that there are only 103056 valid tokens for DeBERTa-v3 and 72513 valid tokens for Llama3-8B-Instruct. 

\begin{table}[h!]
\centering
\resizebox{0.49\textwidth}{!}{%

\begin{tabular}{|l|l|}
\hline
\textbf{Parameter}                        & \textbf{Value}                  \\ \hline
\textbf{Data Source}                      & GLUE training dataset \\ \hline
\textbf{\# of Clusters $k$}                   & 15                             \\ \hline
\textbf{Dimension}                        & 1024                           \\ \hline
\textbf{\# Valid token}                   & 103056                         \\ \hline
\textbf{\# tokens less than 15 senses}     & 50995 (49.48\%)                \\ \hline
\textbf{\# tokens with 15 senses}    & 52061 (50.52\%)                \\ \hline
\textbf{CPU Memory Cost}                  & 2.1 GB                        \\ \hline
\end{tabular}
}
\caption{Sense Dictionary for DeBERTa-v3-large.}
\label{tab:glue_wikitext_stats}
\end{table}

\begin{table}[h!]
\centering
\resizebox{0.49\textwidth}{!}{%
\begin{tabular}{|l|l|}
\hline
\textbf{Parameter}                        & \textbf{Value}                  \\ \hline
\textbf{Data Source}                      & MTEB classification training dataset \\ \hline
\textbf{MCL Inflation}                    & 1.65                           \\ \hline
\textbf{MCL expansion}                    & 2                              \\ \hline
\textbf{Dimension}                        & 4096                           \\ \hline
\textbf{\# Valid token}                   & 72513                          \\ \hline

\textbf{\# tokens less or equal to 15 senses}     & 57888 (79.83\%)                \\ \hline
\textbf{\# tokens more than 15 senses}    & 14625 (20.17\%)                \\ \hline
              
\textbf{CPU Memory Cost}                  & 8.4 GB                          \\ \hline
\end{tabular}
}
\caption{Sense Dictionary for LLaMA-3-8B-Instruct.}
\label{tab:llama_mteb_stats}
\end{table}

\begin{table*}[ht!]
\centering
\resizebox{\textwidth}{!}{%
\begin{tabular}{|l|c|c|c|c|}
\hline
 & \textbf{DeBERTa-v3-large} & \textbf{DeBERTa-v3-xsmall} & \textbf{Our student model: SKD} & \textbf{Distil-DeBERTa} \\ \hline
\# of Parameters & 300M & 22M & 22M & 22M \\ \hline
GPU Memory for Parameters & 600MB & 44MB & 44MB & 44MB \\ \hline
\# of Layers & 24 & 6 & 6 & 6 \\ \hline
Hidden Dim & 1024 & 384 & 384 & 384 \\ \hline
Embedding Size & 1024 & 384 & 384 & 384 \\ \hline
\# of Vocab & 128,100 & 128,100 & 128,100 & 128,100 \\ \hline
Activation Function & GeLU & GeLU & GeLU & GeLU \\ \hline
Precision & FP16 & FP16 & FP16 & FP16 \\ \hline
\end{tabular}
}
\caption{Parameters of Model Architecture for Knowledge Distillation on DeBERTa-v3.}
\label{tab:app_modelsize_deberta}
\end{table*}

\begin{table*}[ht!]
\centering
\resizebox{0.7\textwidth}{!}{%
\begin{tabular}{|l|c|c|}
\hline
 & \textbf{Llama3-8B-Instruct} & \textbf{Our student model: SKD} \\ \hline
\# of Parameters & 7.5B & 1.4B \\ \hline
GPU Memory for Parameters & 15 GB & 2.7GB \\ \hline
\# of Layers & 32 & 4 \\ \hline
Hidden Dim & 4096 & 4096 \\ \hline
Embedding Size & 4096 & 4096 \\ \hline
\# of Vocab & 128,256 & 128,256 \\ \hline
Activation Function & SiLU & SiLU \\ \hline
Precision & BF16 & BF16 \\ \hline
\end{tabular}
}
\caption{Parameters of Model Architecture for Knowledge Distillation on Llama-3-8b-Instruct.}
\label{tab:app_modelsize_deberta_llama}
\end{table*}

For DeBERTa-v3-large we always use $k=15$ as the number of clusters or senses, given the relative smaller size of the GLUE training data. Nevertheless, certain tokens appear only once or twice in the training dataset, and many have limited senses, resulting in many tokens having fewer than 5 senses, as we can see in Table~\ref{tab:glue_wikitext_stats}.
For Llama3-8B-Instruct model, we employ the MCL~\cite{van2008graph} algorithm to determine approximately how many clusters the data may contain. The MCL algorithm does not need the number of clusters as input, but rather uses the inflation parameter (we use a value of $1.65$) to determine the clusters (smaller values result in fewer clusters and larger ones in more clusters). If the number of MCL clusters is larger than 900, we set $k$ to that number multiplied by 0.4 and then use the K-means algorithm to find the final set of $k$ clusters, since this indicates that the token is a generic one and may have many possible senses (e.g., the suffix `ly', which can convert many different verbs to adverbs, and so on). On the other hand, if MCL determines that there are fewer than 900 clusters, we set $k$ to that number multiplied by 0.1. This approach was found to empirically give better results, as described below in Sec.~\ref{sec:k_ablation}.
It is important to note that for our experiments with the student model, if a token is not found in our sense dictionary during inference (i.e., when the token was not seen in the training data), we keep the contextual embedding generated by the last hidden layer from the student model.

As we can observe in Tables~\ref{tab:glue_wikitext_stats} and \ref{tab:llama_mteb_stats}, the sense dictionary occupies 2.1GB space for DeBERTa-v3-large and 8.4GB space for Llama3-8B-Instruct. Nevertheless, the full sense dictionary is 
kept in the CPU memory, and only the subset of token senses active in the current context window needs to be loaded into GPU memory. 
For the large Llama3-8B-Instruct model, which has a context size of 8192 and hidden dimensionality of 4096, this requires at most 670MB GPU space, with 16-bit precision, and $k=15$ senses per token (in practice the space is even smaller due to repeated tokens, and many tokens having fewer than $k$ senses). Likewise, for the smaller DeBERTa model, the active GPU memory for the sense embeddings is about 15MB (using context size 512, hidden dimensionality 1024, 16-bit precision and $k=15$ senses).

\section{Knowledge Distillation: Model Parameters}

Table \ref{tab:app_modelsize_deberta} presents the architectural specifications of the DeBERTa-series~\cite{he2021debertav3} models used in our experiments. Notably, our knowledge-distilled models follow the same architecture as the DeBERTa-v3-xsmall model.
Additionally, Table \ref{tab:app_modelsize_deberta_llama} provides the architectural details of the Llama3-8B-Instruct model~\cite{touvron2023llama} in our experiments. The only modification we make to the student model is reducing the number of layers from 32 to 4. Note that we use reduced-but FP16 training for the DeBERTa models and BF16 for the Llama3-8B model. While both use 16-bit precision, FP16 has a smaller range but higher precision since it uses a 10-bit mantissa, whereas BF16 has a wider range but lower precision since it uses 7-bit mantissa.

In terms of space savings for the student model, we can observe that compared to DeBERTa-v3-large that has 300M parameters and requires about 600MB GPU memory, our SKD student model requires only 22M parameters (due to reduced hidden dimensionality) and 44MB of GPU memory, which matches the DeBERTa-v3-xsmall and the Distil-DeBERTa baselines. For the much larger Llama3-8B model, we can see more significant space benefits. The full model has 7.5B parameters, and requires 15GB GPU memory, whereas our SKD model has 1.4B parameters and requires only 2.7GB GPU memory. 
Also, as noted above, while the sense dictionary occupies 2.1GB space for DeBERTa-v3 and 8.4GB space for Llama3-8B, only about 10MB GPU space is required for DeBERTa and at most 670MB for Llama, to maintain the active senses.

\begin{table}[h!]
\centering
\resizebox{0.49\textwidth}{!}{%
\begin{tabular}{|l|c|c|}
\hline
\textbf{Hyperparameter} & \textbf{DeBERTa-v3} & \textbf{Llama3-8B} \\ \hline
Learning Rate           & {5e-4,1e-3}              & 4e-5         \\ \hline
Batch Size           &  32             & 8          \\ \hline
Sense Emb Similarity           &  Dot Product         & Dot Product       \\ \hline
Precision               & FP16                & BF16              \\ \hline
Epoch                   & \{15, 40\}           & \{2,5\}                \\ \hline
\# of GPU                   & 6       & 6             \\ \hline
Each GPU Memory                  & 32GB        & 32 GB    \\ \hline 
Running Time                  & Up to 6 Hours        & Up to 6 Hours   \\  \hline 
\end{tabular}
}
\caption{Hyperparameters for Knowledge Distillation on DeBERTa-v3-large and Llama3-8B-Instruct.}
\label{tab:app_experimental}
\end{table}

Table \ref{tab:app_experimental} lists the hyperparameters we use to train the student model during the knowledge distillation process on DeBERTa-v3-large and Llama3-8B-Instruct. For both models, we perform a grid search over the learning rates ranging from 1e-5 to 5e-3 (incrementing by 1e-5) and select the best-performing value. 
For the DeBERTa model, we train each configuration for at least 15 epochs and at most 40 epochs. For the Llama model, we train each configuration for at least 2 epochs and at most 5 epochs.

\section{Sense-based Word Similarity Task}
\label{sec:wsd}
\begin{table}[ht]
\centering
\resizebox{0.48\textwidth}{!}{%

\begin{tabular}{|l|c|c|c|c|c|c|}
\hline
\textbf{Benchmark} & \textbf{Word2Vec} & \textbf{Dict2Vec} & \textbf{DRG2Vec} & \textbf{BERT-SD} & \textbf{Flan-T5-SD} \\ \hline
MEN-TR-3K & 0.612 & 0.688 & 0.721 & 0.712 & 0.715 \\  
MTurk-287 & 0.577 & 0.568 & 0.558 & 0.558 & 0.618 \\ 
MTurk-771 & 0.540 & 0.609 & 0.641 & 0.633 & 0.705 \\ 
RG-65 & 0.617 & 0.814 & 0.845 & 0.819 & 0.848 \\ 
WS-353-REL & 0.548 & 0.579 & 0.605 & 0.671 & 0.584 \\ 
YP-130 & 0.257 & 0.528 & 0.610 & 0.666 & 0.683 \\ 
SimLex999 & 0.338 & 0.444 & 0.476 & 0.499 & 0.555 \\
SimVerb-3500 & 0.190 & 0.379 & 0.425 & 0.425 & 0.439 \\
WS-353-SIM & 0.679 & 0.696 & 0.728 & 0.794 & 0.799 \\
Card-660 & 0.234 & 0.348 & 0.513 & 0.364 & 0.394 \\
RW-STANFORD & 0.398 & 0.476 & 0.482 & 0.364 & 0.425 \\ 
MC-30 & 0.682 & 0.748 & 0.738 & 0.884 & 0.912 \\ 
WS-353-ALL & 0.626 & 0.682 & 0.699 & 0.727 & 0.739 \\ \hline
\textbf{Average} & \textbf{0.416} & \textbf{0.529} & \textbf{0.558} & \textbf{0.549} & \textbf{0.591} \\ \hline
\end{tabular}
}
\caption{Word Similarity Test: Spearman Rank Correlation; higher is better.}
\label{tab:wordsimilarity}
\end{table}

We further test the effectiveness of our sense embeddings on the word similarity task, which comprises a list of word pairs with human-provided similarity scores. 
Since the input is just a pair of words, there is no context that the LLMs can leverage. Instead, we first use the English Wikipedia dump~\cite{wikipedia_dump_20240320} as the corpus to construct the word-level sense dictionary, by averaging the embeddings of all of the tokens of that word. Given two words $w_1$ and $w_2$, we compute the dot product between all $k$ senses of each word, and record the maximum similarity. That is, let $\mathbf{s}_1^{(i)}$ and $\mathbf{s}_2^{(j)}$ denote the set of senses for $w_1$ and $w_2$, respectively; their similarity is given as:
$\max_{i,j}\{ \langle \mathbf{s}_1^{(i)}, \mathbf{s}_2^{(j)} \rangle \}$ for $i,j=1,...,k$.
After obtaining the scores for all word pairs, we rank them from highest to lowest.
We then compute the Spearman rank correlation with ground-truth human similarity scores, where a higher correlation indicates that the model’s embeddings capture more semantic information, and closely matches the ground truth. 

We evaluate on standard word similarity benchmarks, including
Card-660~\citep{pilehvar2018card}, MC-30~\citep{miller1991contextual}, MEN-TR-3K~\citep{bruni2014multimodal}, MTurk-287~\citep{radinsky2011word}, MTurk-771~\citep{halawi2012large}, RG-65~\citep{rubenstein1965contextual}, RW-STANFORD~\citep{luong2013better}, SimLex-999~\citep{hill2015simlex}, SimVerb-3500~\citep{gerz2016simverb}, WS-353-ALL~\citep{finkelstein2001placing}, WS-353-REL~\citep{finkelstein2001placing}, WS-353-SIM~\citep{finkelstein2001placing}, and YP-130~\citep{yang2006verb}. 
These benchmarks provide a list of word pairs with human-evaluated similarity scores, where a high score indicates similarity and a low score suggests little or no relation between the words. Since these are word-based benchmarks, we represent each word by averaging the token embeddings to form the word embedding. For our experiments, we use $5\%$ of the \citet{wikipedia_dump_20240320} as our resource and collect the last hidden layer of the encoder, extracting up to 8,000 embeddings per word. We then cluster each word embedding into the default $k=5$ clusters using K-means.


In Table \ref{tab:wordsimilarity}, we compare the sense embeddings
derived from BERT~\cite{devlin2018bert}, denoted BERT-SD, and Flan-T5-xxl~\cite{chung2024scaling}, denoted Flan-T5-SD, with discrete embeddings from Word2Vec~\cite{mikolov2013efficient}, Dict2Vec~\cite{shu2020drg2vec} and DRG2Vec~\cite{tissier2017dict2vec}.
Flan-T5 is an encoder-decoder model, but we utilize only the encoder to create the sense dictionary.
We can see that the sense-based (centroids derived from) BERT and Flan-T5 perform competitively compared to traditional discrete embeddings. Additionally, the larger Flan-T5-SD
model has significantly better performance, underscoring the potential of larger models to enhance embedding quality.

\section{Ablation Study: Number of Layers in Knowledge Distillation}
\begin{figure}[!ht]
\centering
\includegraphics[width=0.45\textwidth]{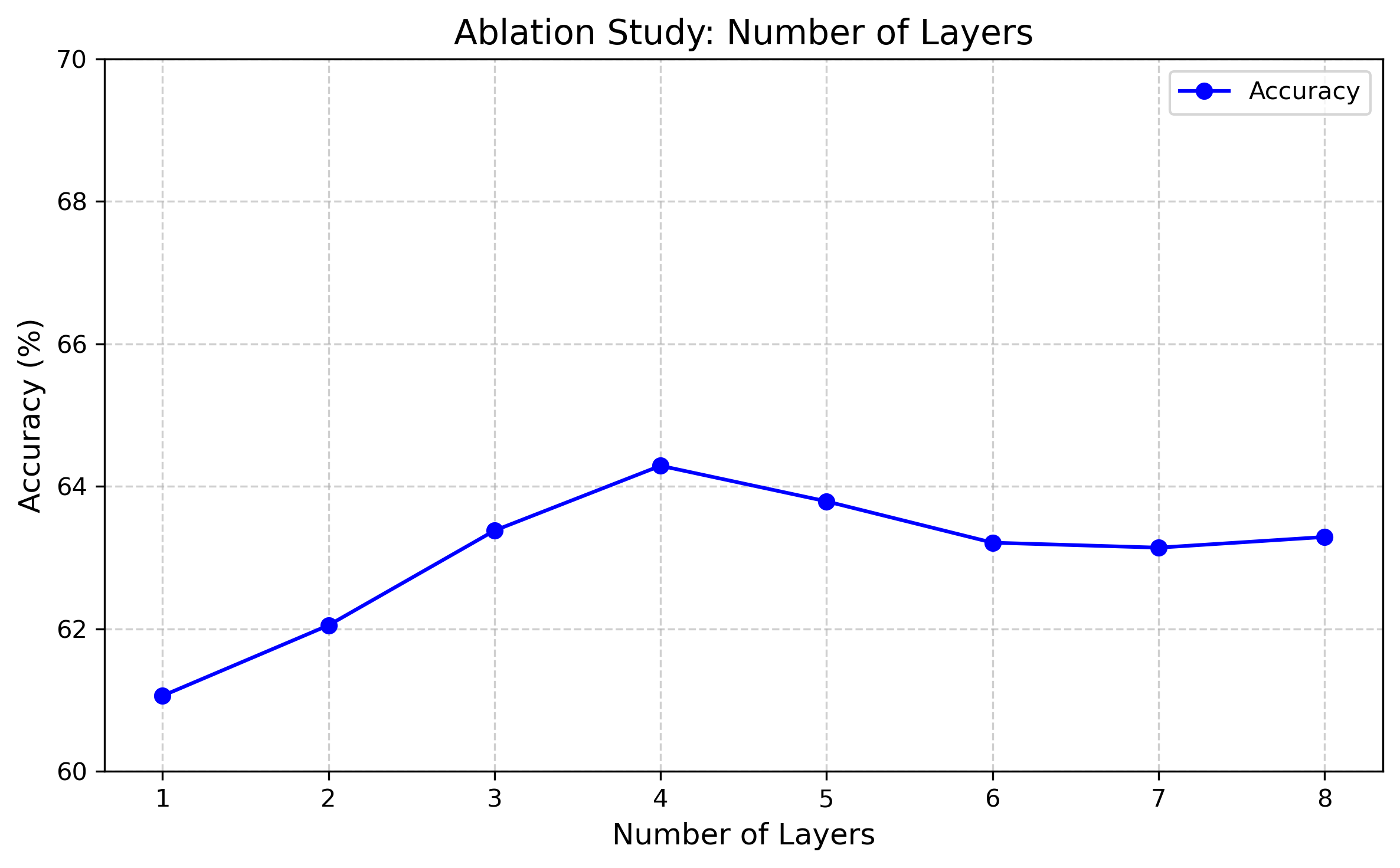} %
\caption{Ablation Study: Number of Layers.}
\label{fig:layers_study}
\end{figure}

We conduct an ablation study to investigate the impact of the number of layers in the SKD student model in our sense-based knowledge distillation approach.  Fig.~\ref{fig:layers_study} illustrates the relationship between the number of layers and the model's performance on the FrenkEn classification dataset of MTEB~\cite{muennighoff2022mteb} with Llama3-8B-Instruct as the teacher model. We evaluate the accuracy of the student model with varying numbers of layers, ranging from 1 to 8. While the accuracy improves as the number of layers increases from 1 to 4, further increases in the number of layers show a plateau or even a slight decline in performance. This suggests that deeper models may not necessarily enhance the capacity of the student model to generalize effectively. Based on these findings, we selected 4 layers as the default configuration for our student model, balancing accuracy and simplicity. 

\section{Ablation Study: Hidden Layer Dimension in Knowledge Distillation}
\begin{figure}[!ht]
\centering
\includegraphics[width=0.45\textwidth]{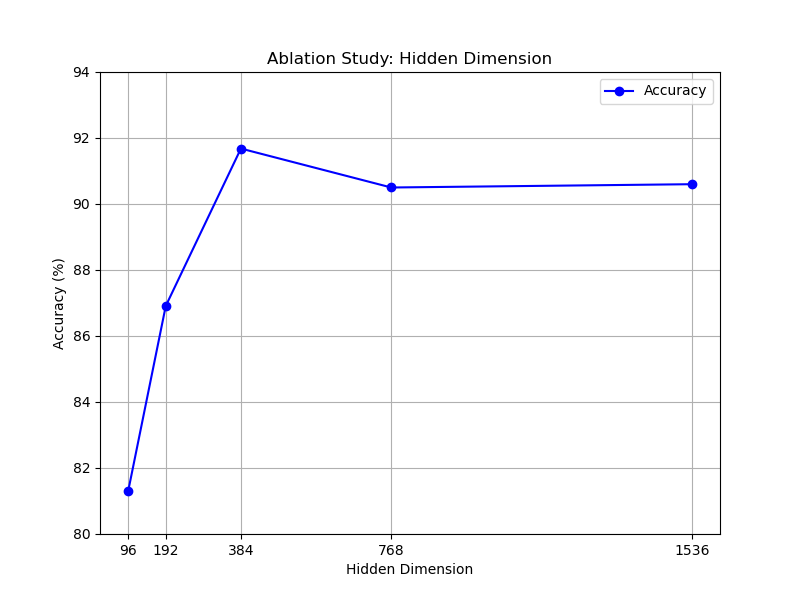} %
\caption{Ablation Study: Hidden Layer Dimension.}
\label{fig:hidden_dim}
\end{figure}

We perform an ablation study to explore the impact of the hidden layer dimension in the SKD student model.
As shown in Fig.~\ref{fig:hidden_dim}, we explore the relationship between hidden layer dimension and performance on the STS-B classification task from the GLUE benchmark. The student model is evaluated across a range of hidden dimensions—96, 192, 384, 768, and 1536—where each value doubles the previous one. The results indicate that very small dimensions (e.g., 96 or 192) lead to significantly lower accuracy. While increasing the dimension improves performance initially, further increases beyond 384 yield diminishing returns and incur higher computational and memory costs. Notably, the best performance is achieved at a dimension of 384, which aligns with the default configuration of DeBERTa-v3-xsmall. Based on this observation, we choose 384 as the hidden layer dimension in our model.


\section{Ablation Study: Number of Clusters $k$}
\label{sec:k_ablation}

\begin{table}[h!]
\centering
\resizebox{0.4\textwidth}{!}{%
\begin{tabular}{|l|c|c|}
\hline
\textbf{Models} & \textbf{CoLA} & \textbf{SST-2} \\ \hline
Deberta-v3-large (base) & 75.72 & 96.21 \\ \hline\hline
$k=1$ & 14.57 & 59.97 \\ \hline
$k=5$ & 72.19 & 96.10 \\ \hline
$k=10$ & 72.85 & 96.10 \\ \hline
$k=15$ & 73.13 & 96.21 \\ \hline
$k=20$ & 73.05 & 95.87 \\ \hline
$k=25$ & 72.66 & 95.98 \\ \hline
$k=30$ & 72.00 & 96.21 \\ \hline
\end{tabular}
}
\caption{Encoder LLM Replacement Test for DeBERTa-v3-large with different number of clusters $k$.}
\label{tab:app_llm_kcluster}
\end{table}

As noted above, for the sense dictionary for DeBERTa-v3-large, which is an encoder LLM, we use a default value of $k=15$ clusters or senses per token (however, many tokens have fewer senses as noted in Table~\ref{tab:glue_wikitext_stats}).
Table \ref{tab:app_llm_kcluster} shows the effect of varying $k$ for the LLM replacement test
for the DeBERTa-v3-large model \cite{he2021debertav3}, using the CoLA and SST-2 datasets from the GLUE benchmark \cite{wang2019glue}. When $k=1$, each token has only one sense embedding, which eliminates the need to train a student model since the student model can select the single available embedding. Thus, the low performance across both datasets when $k=1$ indicates that a single-sense embedding is insufficient. Additionally, the results show that larger $k$ values do not yield significant improvements. Therefore, we choose $k=15$ for the knowledge distillation of DeBERTa models. 

\begin{table}[h!]
\centering
\begin{tabular}{|l|c|}
\hline
\textbf{Model} & \textbf{Score} \\ \hline
Llama3-8B (base) & 86.68 \\ \hline\hline
$k=10$ & 78.63 \\ \hline
$k=20$ & 80.49 \\ \hline
$k=40$ & 80.75 \\ \hline
$k=80$ & 80.69 \\ \hline
$k=100$ & 80.64 \\ \hline
$k=150$ & 80.98 \\ \hline\hline
MCL with 0.1 & 82.35 \\ \hline
MCL with 0.2 &  81.56 \\ \hline
MCL with 0.3 & 84.39 \\ \hline
MCL with 0.4 & 85.34 \\ \hline\hline
MCL with 0.05/0.4 & 84.45 \\ \hline
MCL with 0.1/0.4 & 85.26 \\  \hline
\end{tabular}
\caption{Decoder LLM replacement test with Llama3-8B-Instruct with different $k$, and with and without MCL.}
\label{tab:ab_kmeans_mcl}
\end{table}

Table~\ref{tab:ab_kmeans_mcl} presents the impact of the number of clusters $k$ and various clustering strategies on the performance of the Llama3-8B-Instruct decoder model \cite{touvron2023llama} for the FinancialPhrasebank classification task from the MTEB benchmark \cite{muennighoff2022mteb}. We begin by applying K-means clustering with different values of $k$. Although performance increases up to $k=150$, the improvements beyond $k=20$ are marginal.

However, we observed that different tokens may benefit from different numbers of clusters, as some tokens capture a wider range of semantic meanings than others. To accommodate this variability, we employ the MCL clustering method \cite{van2008graph}, which can automatically estimate the appropriate number of clusters. We begin by conducting a grid search over the inflation (1.05 to 1.95) and expansion (2 to 5) parameters. In practice, we find these parameters to be highly sensitive—even small changes can lead to significant drops in performance. Finally, we select inflation = 1.65 and expansion = 2, which yield performance comparable to that of LLaMA3-8B-Instruct. However, MCL tends to produce an excessive number of clusters, resulting in an impractically large sense dictionary. To mitigate this, we retain only a fraction of the MCL-generated clusters ( scaling coefficient) and subsequently apply K-means using the reduced number of clusters determined by MCL.
In the second part of Table~\ref{tab:ab_kmeans_mcl}, ``MCL with 0.1'' indicates that only 10\% of the clusters are preserved. Notably, MCL consistently outperforms K-means at equivalent $k$ values, with ``MCL with 0.4'' achieving the highest accuracy. However, this comes at the cost of an excessively large dictionary.

To balance accuracy and memory efficiency, we introduce a filtering strategy based on the total number of clusters generated by MCL. Specifically, if the number of clusters is fewer than 900, we retain only 0.05 or 0.1 of them; otherwise, we use a 0.4 fraction. Results from this adaptive filtering approach are shown in the third part of Table~\ref{tab:ab_kmeans_mcl}. Ultimately, we choose the "MCL with 0.1/0.4" setting, which significantly reduces the dictionary size while preserving strong classification performance.

\appendix



\end{document}